# A Data-Driven Approach to Positioning Grab Bars in the Sagittal Plane for Elderly Persons

Roberto Bolli Jr.
*Department of Mechanical Engineering*
*MIT*
Cambridge, USA
rbolli@mit.edu

H. Harry Asada
*Department of Mechanical Engineering*
*MIT*
Cambridge, USA
asada@mit.edu

*Abstract*—The placement of grab bars for elderly users is based largely on ADA building codes and does not reflect the large differences in height, mobility, and muscle power between individual persons. The goal of this study is to see if there are any correlations between an elderly user's preferred handlebar pose and various demographic indicators, self-rated mobility for tasks requiring postural change, and biomechanical markers. For simplicity, we consider only the case where the handlebar is positioned directly in front of the user, as this confines the relevant body kinematics to a 2D sagittal plane. Previous eldercare devices have been constructed to position a handlebar in various poses in space. Our work augments these devices and adds to the body of knowledge by assessing how the handlebar should be positioned based on data on actual elderly people instead of simulations.

## I. Introduction

The elderly population is at an increased risk of falls and injuries, which can lead to hospitalization and a decline in overall health [1]. Handrails and grab bars have been recognized as effective interventions for reducing the risk of falls by providing stability and support while standing, walking, or transferring from one surface to another. Additionally, grab bars can enhance the independence and quality of life of elderly individuals, allowing them to perform daily activities with greater ease and safety [2].

Despite this, there is still a lack of implementation of grab bars in both public and private spaces. It is costly and impractical to put grab bars in every area of the home, so users will typically only have them installed to support high-risk activities such as getting out of the bathtub. This means that there will be many scenarios where there may not be adequate (or any) support for the elderly person. Previous work [3] has sought to solve this issue through the use of a robotic repositionable handlebar that eliminates the need for installing individual grab bars, and has the added benefit of allowing the handlebar to be positioned at any point in space. These new eldercare technologies necessitate the need to assess elderly persons' handlebar placement preference when the potential handlebar location is not constrained by the room layout, as is the case for traditional grab bars.

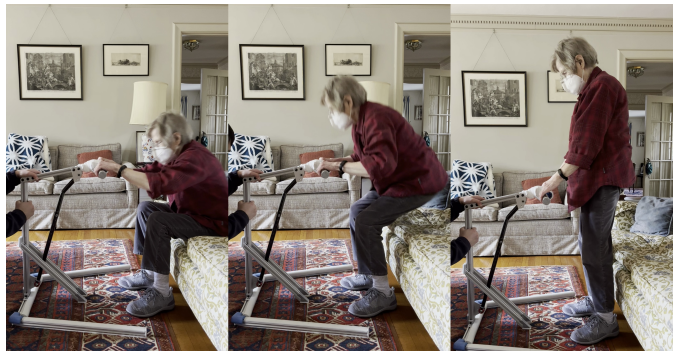

Fig. 1. An elderly subject performing a sit-to-stand transition using the handlebar manipulandum. Photo included with consent.

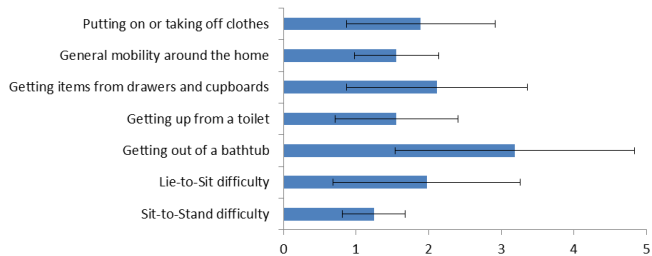

Fig. 2. Respondents' self-rated difficulty for various common postural changes. The black bars represent one standard deviation.

## II. Methodology

The study protocol was reviewed and approved by the MIT IRB committee. Participants were asked to rate the difficulty of performing everyday activities on a scale of 1-5, with 1 being the easiest and 5 being the hardest. Additionally, participants were asked a set of open-ended questions about various demographic factors, difficulty in postural changes, and health status. A handlebar manipulandum (Fig. 1) was presented to each subject, and the height and radial distance of the handlebar were adjusted until each participant felt that it was in the most comfortable location to perform a sit-to-stand (STS) transition. The handlebar was located in front of the user to confine the relevant biomechanics to a 2D sagittal plane. See Appendix 1 for the data on each participant.

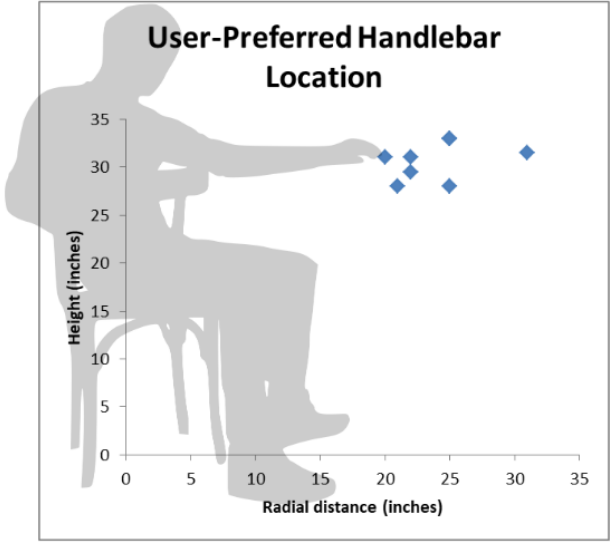

Fig. 3. Spatial distribution of user-preferred handlebar locations. The handlebar was located directly in front of the user, in the sagittal plane. Radial distance was measured from the center of the user's body. The chair height was approximately 18".

## III. RESULTS

### A. Survey Responses

In total, 9 persons over the age of 65 (3 females, 6 males) participated in the study. One participant was unable to perform the sit-to-stand transitions, but answered the questions from the non-experimental section of the study. The respondents were from a database of mentally healthy elderly persons hosted by the MIT AgeLab. The average age was 86.11 ± 9.61 years (1 standard deviation). Participants' average self-rated physical health was 2.5 ± 1.1, on a scale of 1 to 5 with 1 being healthiest. 89% of the participants exercised regularly.

Respondents' self-rated difficulty for various postural changes can be seen in Fig. 2. Of the scenarios, getting out of a bathtub was rated to be the most difficult, which is consistent with the widespread adoption and use of grab bars in the bathroom. Additionally, we asked the elderly persons to name the most difficult postural change that was not part of the scenarios in Fig. 2. The most common responses were picking up heavy objects (x3 respondents), standing from a kneeling position (x2), getting up from sitting on the floor, looking upwards (due to neck pain), carrying groceries, and bending over.

As seen in the spatial distribution of participants' preferred handlebar location in Fig. 3, the radial handlebar distance varied more than the handlebar height, although both varied considerably. This reinforces the notion that a fixed handlebar location is not ideal for individual elderly users.

### B. Data Analysis

Least-squares linear regressions were performed on a relevant subset of the variables asked in the study to see if there was any correlation between each variable and the handlebar height or radial distance from the user. The variables considered were STS difficulty, reaching difficulty, general mobility rating, STS duration without handlebar, age, height, weight, and physical health score. To avoid overfitting, we limited the number of independent variables in the regression to either one variable or two variables and an interaction term. This meant that the regressions took on the form of $y = \beta_0 + \beta_1 x$ or $y = \beta_0 + \beta_1 x_1 + \beta_2 x_2 + \beta_3 x_1 x_2$.

TABLE I
COEFFICIENTS FOR THE LINEAR REGRESSIONS

|  | $\beta_0$ | $\beta_1$ | $\beta_2$ | $\beta_3$ |
| --- | --- | --- | --- | --- |
| 1D Radial Distance | 30.6182 | -4.6909 |  |  |
| 1D Radial Dist. (Func) | 27.4927 | -1.5858 |  |  |
| 2D Radial Distance | -151.1486 | 1.9494 | 1.4899 | -0.0166 |
| 2D Radial Dist. (Func) | -52.3171 | 0.0094 | 23.1483 | -0.0029 |
| 1D Height | 22.3596 | 0.0615 |  |  |
| 1D Height (Func) | 37.9080 | -945.5490 |  |  |
| 2D Height | -259.7438 | 4.5251 | 119.9324 | -1.8894 |
| 2D Height (Func) | 119.469 | -369665 | -15.8633 | 63479.7 |

In both of these cases, we also processed the data through nine elementary functions to see if the relationship could be better explained through a non-linear law. The functions were $\sqrt{(x)}, x^2, \ln(x), e^{x/100}, \frac{1}{x}, \frac{1}{\sqrt{(x)}}, \frac{1}{x^2}, \frac{1}{\ln(x)}, \frac{1}{e^{x/100}}$. Thus, four linear regressions were performed in total. Table 1 shows the coefficients for each of the linear regressions. The independent variables used in the one variable regressions are described in the following paragraph, while the variables used in the two variable regressions are shown in the figures below.

Both of the $R^2$ values for the single variable linear regressions were low (0.45 for the radial distance and 0.59 for the height). The handlebar radial distance correlated best with the general mobility score, while the handlebar height interestingly correlated best with the subjects' weight. The highest $R^2$ value when the data was processed through each of the elementary functions described previously remained virtually the same, increasing to 0.46 and 0.61, respectively, for the square of the general mobility and the inverse of the user weight ($\frac{1}{\text{weight}}$).

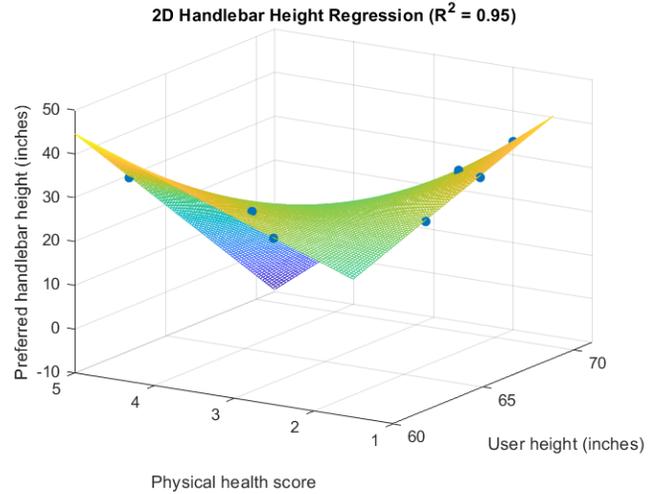

Fig. 4. Mesh visualization of the results of a multiple linear regression on handlebar height. The respondents' preferred heights are shown by the blue dots.

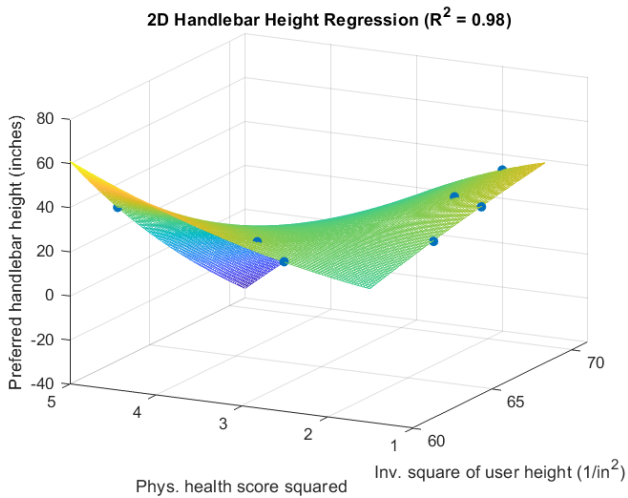

Fig. 5. Multiple linear regression on handlebar height, with the independent variables processed through nonlinear functions.

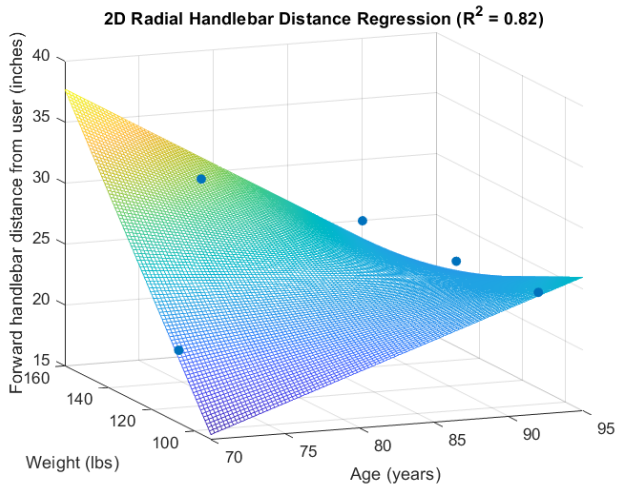

Fig. 6. Multiple linear regression on the handlebar's radial distance from the user, which is the horizontal distance from the body center of mass (CoM) in the sagittal plane.

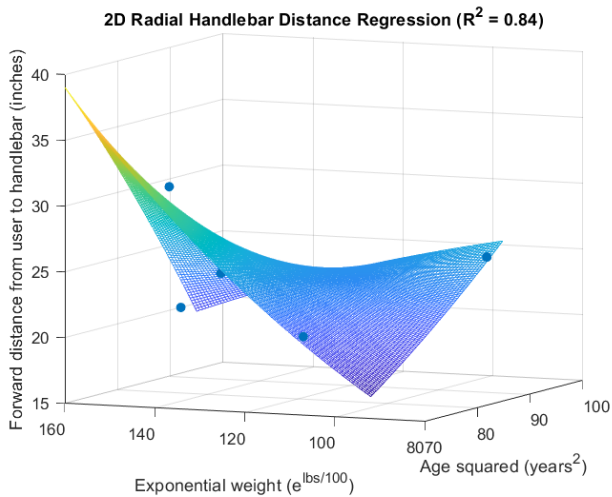

Fig. 7. Multiple linear regression on the handlebar's radial distance, with the independent variables processed through nonlinear functions.

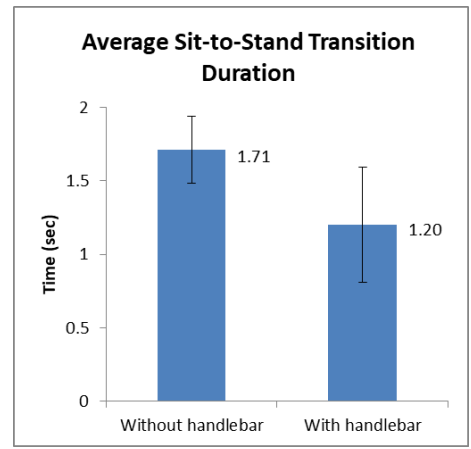

Fig. 8. The average duration for a STS transition decreased by 29.77% with the frontal handlebar. The black bars represent the sample standard deviation. A one-tailed unequal variance (Welch) t test found the decrease to be significant relative to $\alpha = 0.05$, with a p value of 0.0042.

For the linear regressions with two variables, we saw a very strong correlation ($R^2 = 0.98$) between handlebar height, user height and physical health score (Figs. 4 and 5). The radial handlebar distance did not have quite as strong as a correlation, reaching a maximum $R^2$ value of 0.84 with age and weight (Figs. 6 and 7).

## IV. CONCLUSION AND DISCUSSION

It appears that it may be possible to predict an elderly person's preferred handlebar radial distance and height using just four variables: age, weight, height, and self-rated physical health. In addition, six out of eight (75%) respondents said that the handlebar made it easier for them to stand up. The other 25% had no problem standing up without the handlebar. 56% of the respondents said they would consider using a handlebar assistance robot in their home. Finally, we found that there was a statistically significant decrease in the average time to perform a sit-to-stand transition when the elderly persons used the handlebar at their preferred height (Fig. 8).

We recognize that this study has several limitations on the scope and relevance of the data. Future work will involve recruiting more participants and testing other grab bar configurations besides the sagittal plane.

## ACKNOWLEDGMENT

This material is based upon work supported by the National Robotics Initiative Grant No. 2133075 and the National Science Foundation (NSF) Graduate Research Fellowship under Grant No. 2141064.

APPENDIX 1: USER STUDY DATA

| Patient ID | Sit-to-stand (STS) difficulty 1-5; 1 is easiest | Lie-to-sit difficulty | Getting out of bathtub | Getting up from toilet | Getting items from drawers and cupboards | General mobility around the home | Putting on or taking off clothes |
|---|---|---|---|---|---|---|---|
| P1 | 2 | 2.5 | 3 | 2.5 | 2 | 2 | 3.5 |
| P2 | 1 | 1 | 3 | 3 | 1.5 | 1.5 | 2 |
| P3 | 1 | 2.25 | 5 | 1 | 1 | 2 | 1 |
| P4 | 1 | 1 | 2 | 1 | 1 | 1 | 1 |
| P5 | 1.2 | 1 | 1.5 | 1 | 1.5 | 1 | 1 |
| P6 | 1 | 2 | 1 | 1 | 1 | 1 | 1.5 |
| P7 | 2 | 1 | - | 1 | 3 | 2 | 1 |
| P8 | 1 | 2 | 5 | 1 | 4 | 1 | 3 |
| P9 | 1 | 5 | 5 | 2.5 | 4 | 2.5 | 3 |

| Patient ID | Preferred handlebar radial dist. from CoM | Preferred handlebar height from ground | STS duration without handlebar | STS duration with handlebar | Age | Sex | Height | Weight | Self-rated phys. health 1-5; 1 is healthiest |
|---|---|---|---|---|---|---|---|---|---|
| P1 | 21" | 28" | 1.68 sec | 1.65 sec | 70 | F | 5'4" | 107 lbs | 1.5 |
| P2 | 25" | 33" | 1.46 sec | 0.85 sec | 90 | M | 5'11" | 160 lbs | 2 |
| P3 | 20" | 31" | 2.14 sec | 0.8 sec | 92 | F | 5'1" | 160 lbs | 3 |
| P4 | 25" | 33" | 1.6 sec | 0.9 sec | 74 | F | 5'7" | 130 lbs | 1.5 |
| P5 | 25" | 28" | 1.5 sec | 0.91 sec | 92 | F | 5' | 92 lbs | 2.5 |
| P6 | 31" | 31.5" | 1.83 sec | 1.68 sec | 77 | M | 5'8" | 145 lbs | 2 |
| P7 | 22" | 29.5" | 1.89 sec | 1.63 sec | 94 | F | 5'3" | 130 lbs | 5 |
| P8 | 22" | 31" | 1.61 sec | 1.21 sec | 95 | M | 5'8" | 151 lbs | 2 |
| P9 | - | - | - | - | 91 | F | 5'2" | 121 lbs | 3 |

| Patient ID | Exercises Regularly? | Experiencing any muscle aches or pain? | Any vertigo or motion sickness? | Left or right handed? |
|---|---|---|---|---|
| P1 | Yes (10,000 steps/day, lifts weights 3x week, weekly stretches and pickleball) | Shoulder pain, cervical stenosis | No | Right |
| P2 | Yes (1.5-2 miles of walking a day) | No | Some lightheadedness after standing up after sitting for a long time | Right |
| P3 | Yes (once a week in the morning) | Yes, needing to sit frequently (spinal stenosis) | Yes, in the past | Left |
| P4 | Yes | No | No (only on boats) | Right |
| P5 | Yes | Some (rheumatoid arthritis) | No | Right |
| P6 | Yes (lots of biking, 4-5 days a week) | Occasional discomfort | No | Right |
| P7 | No | No | Lightheadedness during postural change | Right |
| P8 | Exercising in the pool, Tai Chi, 500 steps/day | Not really | Postural dizziness | Right |
| P9 | Exercise in the pool | Chronic pain | Postural dizziness | Right |